\documentclass[runningheads]{llncs}

\usepackage[T1]{fontenc}
\usepackage{amsmath,amssymb}
\usepackage{float}
\usepackage{graphicx}
\usepackage{url}
\usepackage{color}
\usepackage{marvosym}
\usepackage[hidelinks]{hyperref}

\begin{document}

\title{GeoCueFormer: Geometry-Guided Wavelet Representation and Prediction-Cued Dual-Stage Decoder for Underwater Semantic Segmentation}
\titlerunning{GeoCueFormer}
\author{
Xian Wu\inst{1} \and
Xinjin Li\inst{2} \and
Yiliu Xu\inst{3} \and
Yining Liu\inst{4} \and
Yong Jiang\inst{1}\textsuperscript{\Letter}
}

\authorrunning{X. Wu et al.}

\institute{
Southwest University of Science and Technology, Mianyang, China\\
\email{wuxian@mails.swust.edu.cn, jiang\_yong@swust.edu.cn}
\and
Columbia University, New York, United States
\and
Carnegie Mellon University, Pittsburgh, United States
\and
University of California, Berkeley, CA, United States
}

\maketitle

\begin{abstract}

Underwater semantic segmentation is essential for marine ecosystem monitoring, yet remains challenging due to severe visual degradation. Light absorption and scattering often lead to color shifts, low contrast, and blurred boundaries, making shallow detail features unreliable. Existing underwater segmentation methods improve RGB feature aggregation or boundary prediction, but still lack an explicit mechanism to distinguish structure-related details from degradation-induced responses. To address this limitation, we propose GeoCueFormer, a lightweight framework that combines geometry-constrained frequency enhancement with prediction-cued refinement.
GeoCueFormer performs stage-specific wavelet enhancement on hierarchical encoder features to complement shallow boundary details while preserving deep structural semantics. A depth-derived spatial gate constrains shallow frequency enhancement toward geometry-consistent regions, and a prediction-cued dual-stage decoder further refines ambiguous high-resolution features.GeoCueFormer obtains 82.23\% and 73.04\% mIoU on SUIM and DUT, respectively.Under comparable model complexity and standard benchmark settings on SUIM and DUT, it achieves SOTA performance while maintaining a favorable accuracy-complexity trade-off.
These results show that distinguishing structural details from degradation-induced interference is more effective for underwater segmentation.


\end{abstract}

\begin{figure}[t]
    \centering
    \includegraphics[width=0.95\textwidth]{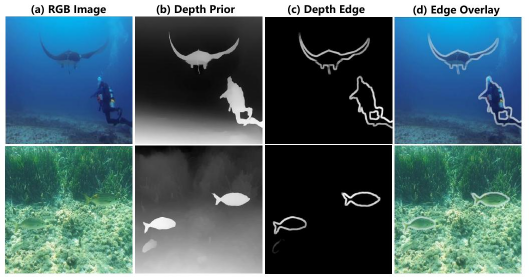}
    \caption{Motivation of depth-guided geometric cues.}
    \label{fig:motivation}
\end{figure}

\section{Introduction}

Underwater semantic segmentation assigns pixel-level labels to underwater scenes and supports applications such as marine ecosystem monitoring and underwater robotic perception. However, underwater images are often degraded by light absorption, scattering, turbidity, and low illumination, leading to color shifts, reduced contrast, and blurred boundaries. These degradations make dense prediction difficult, especially for boundary-sensitive and low-contrast regions.

A key challenge is that shallow high-resolution features are both necessary and unreliable. They preserve local boundaries and fine details, but these responses can be mixed with background textures, scattering artifacts, and illumination variations in degraded underwater images. Deep features provide more stable semantics but lack spatial precision. Thus, directly enhancing or fusing shallow details may introduce degradation-induced interference together with useful structural cues.

Existing underwater segmentation methods mainly improve RGB feature aggregation, high-resolution representation, attention recalibration, or boundary prediction. While these designs enhance contextual modeling and boundary awareness, they provide limited explicit modeling of whether fine details correspond to real structures or degradation-induced responses. This motivates our core question: how can a lightweight segmentation model exploit structure-related details while suppressing degradation interference?

We introduce \mbox{GeoCueFormer}, a lightweight SegFormer-B0-based framework that combines frequency separation, geometric reliability, and prediction-guided refinement. Geometry-Guided Stage-Specific Wavelet Enhancement selects different wavelet bands for different encoder stages and uses a depth-derived spatial gate to constrain shallow frequency injection. Prediction-Cued Dual-Stage Decoder further uses the first-stage prediction to refine ambiguous high-resolution features before final decoding. This design avoids uniform frequency amplification and keeps deep semantic features stable.

Our contributions are summarized as follows:
\begin{itemize}
\item We formulate degraded underwater segmentation as a reliability-aware detail modeling problem, aiming to separate structure-related cues from degradation-induced interference.
\item We propose \mbox{GeoCueFormer}, a lightweight framework that combines geometry-guided stage-specific wavelet enhancement with prediction-cued dual-stage refinement.
\item Experiments on SUIM and DUT demonstrate that \mbox{GeoCueFormer} achieves SOTA-level performance under comparable model complexity and standard benchmark settings, while ablations and sensitivity analyses further validate the effectiveness of the proposed design.
\end{itemize}

To facilitate reproducibility, the source code is publicly available at \url{https://github.com/xianw-u/GeoCueformer}.

\section{Related Work}

\subsection{Underwater Semantic Segmentation}

Underwater images are often degraded by absorption, scattering, and turbidity, leading to color casts, low contrast, blurred boundaries, and detail loss. These degradations weaken pixel-level discrimination and make underwater semantic segmentation challenging. To address this issue, recent methods have introduced degradation-aware network designs. UISS-Net~\cite{he2024uissnet} enhances boundary-region features to reduce pixel-level confusion near object edges. UWSegFormer~\cite{zuo2025uwsegformer} adapts SegFormer to low-quality underwater images through image-quality-aware attention, multi-scale aggregation, and edge-aware supervision. UHRS-Net~\cite{zhou2025uhrsnet} emphasizes high-resolution representation to preserve local details and boundary structures. These studies highlight the importance of modeling degraded boundaries and insufficient feature responses, while reliable structural cues remain less explored.

\subsection{Transformer-Based Semantic Segmentation}

Transformers have been widely adopted in dense prediction tasks due to their ability to model long-range dependencies beyond local convolutions. Swin Transformer~\cite{liu2021swin} introduces shifted-window attention for efficient hierarchical representation, while SegFormer~\cite{xie2021segformer} combines a hierarchical Transformer encoder with a lightweight MLP decoder. SegFormer's hierarchical features are well suited to underwater semantic segmentation, as shallow features preserve boundary details while deep features capture semantic context. However, underwater degradations can still weaken the reliability of hierarchical features, making it important to enhance structural information while suppressing degradation-induced interference.

\subsection{Multi-scale feature fusion}

High-level features provide rich semantics but limited spatial resolution, whereas low-level features preserve fine details with weaker semantics, making multi-scale fusion essential for integrating their complementary strengths. However, the semantic gap across hierarchical features makes direct fusion insufficient for fully exploiting cross-scale complementarity.

Representative methods such as DeepLabv3+~\cite{chen2018encoder} and HRNet~\cite{wang2021deep} improve segmentation by integrating high-level semantic context with low-level spatial details or maintaining high-resolution representations.

These methods demonstrate the effectiveness of cross-level fusion. However, in degraded underwater scenes, low-level details can be unreliable, and direct fusion may introduce noise along with useful boundary cues. Therefore, underwater semantic segmentation requires hierarchical feature integration with reliability-aware and scale-selective enhancement.

\section{Method}

\begin{figure}[t]
    \centering
    \includegraphics[width=\textwidth]{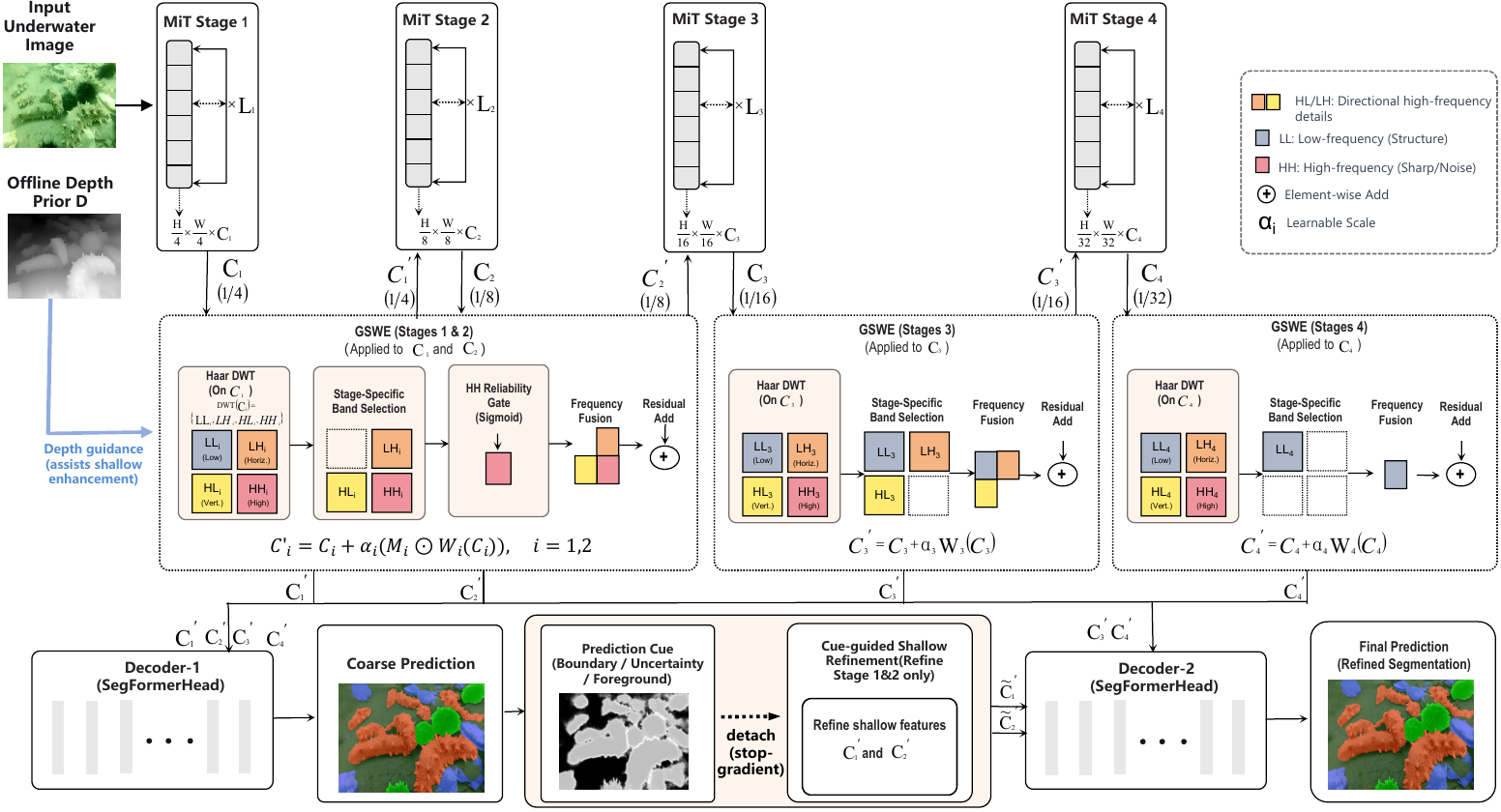}
    \caption{Overall architecture of GeoCueFormer.}
    \label{fig:overall}
\end{figure}

\mbox{GeoCueFormer} is a lightweight framework for degraded underwater semantic segmentation. As shown in Fig.~\ref{fig:overall}, it uses SegFormer-B0 to extract hierarchical features from an underwater RGB image $I$ and introduces a precomputed single-channel depth prior $D$ as auxiliary geometric guidance. The encoded features are enhanced by Geometry-Guided Stage-Specific Wavelet Enhancement (GSWE) and then decoded by the Prediction-Cued Dual-Stage Decoder (PCDD) to produce the final segmentation result.

\noindent\textbf{Depth-prior generation.}
We generate monocular depth maps using the pretrained Depth Anything V2 Small model~\cite{yang2024depthanythingv2}, without fine-tuning or joint optimization with GeoCueFormer. All depth maps are generated offline before segmentation training and evaluation. Each depth prediction is resized to the original image resolution using bicubic interpolation and independently normalized to $[0,1]$, resulting in a single-channel depth prior for each RGB image.

\begin{figure}[!b]
    \centering
    \includegraphics[width=\textwidth]{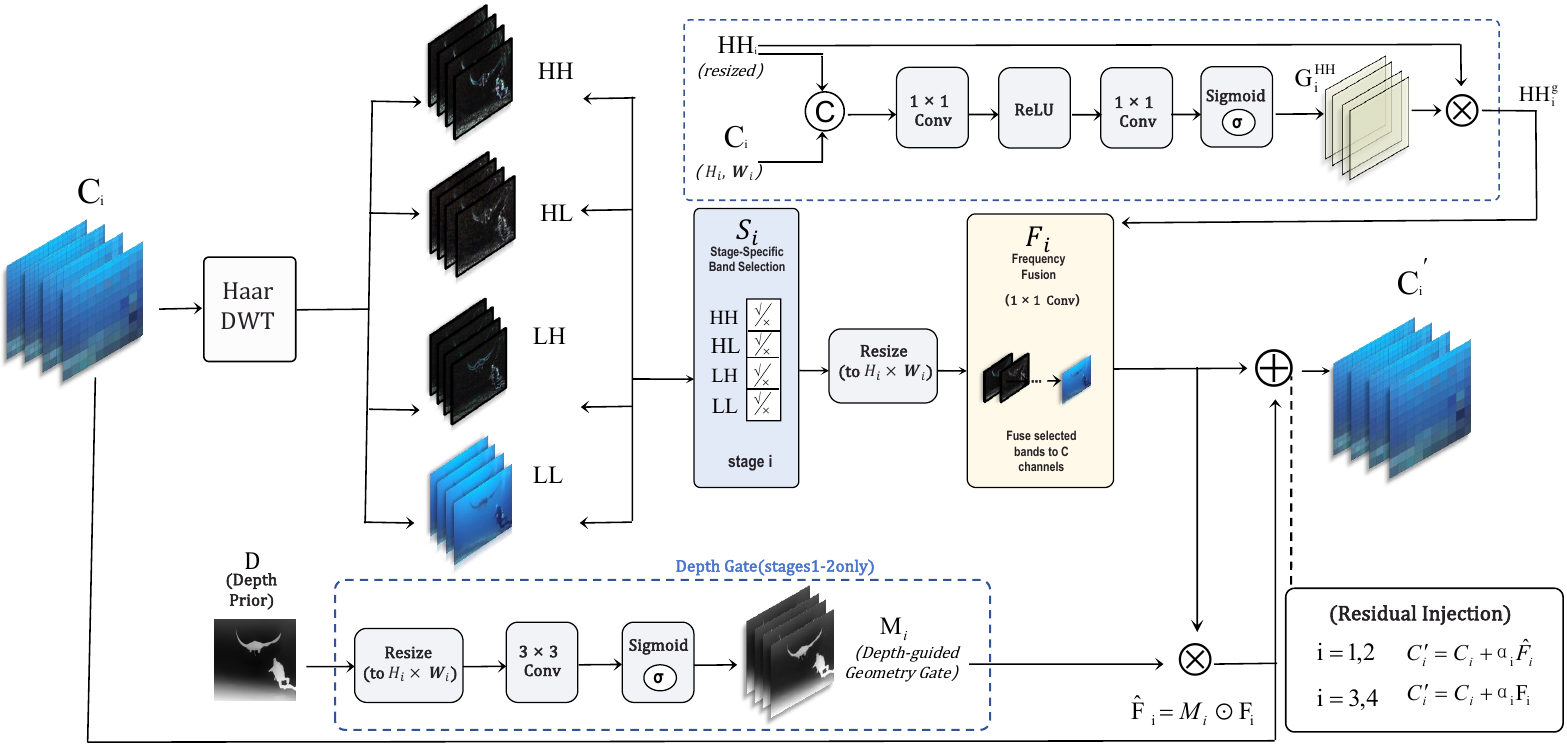}
    \caption{Stage-Specific Wavelet Representation with Geometry-Guided Gating}
    \label{fig:wavelet_gate}
\end{figure}

\subsection{Geometry-Guided Stage-Specific Wavelet Enhancement}

Different encoding stages of SegFormer-B0 play distinct roles: shallow features are sensitive to high-frequency details, whereas deep features require stable structural semantics. Therefore, GeoCueFormer adopts stage-specific wavelet enhancement instead of a uniform frequency enhancement strategy.

As shown in Fig.~\ref{fig:overall}, given the output feature $C_i\in\mathbb{R}^{d_i\times H_i\times W_i}$ from the $i$-th encoding stage, we apply the Haar discrete wavelet transform $\mathcal{W}(\cdot)$ to obtain one low-frequency component and three directional high-frequency components:
\begin{equation}
\{LL_i,LH_i,HL_i,HH_i\}=\mathcal{W}(C_i).
\end{equation}
The Haar wavelet is used for its simplicity and efficiency, where $LL_i$ mainly represents low-frequency structural information, and $LH_i$, $HL_i$, and $HH_i$ encode directional high-frequency responses related to boundaries and textures. Rather than reconstructing the original features using inverse wavelet transform, we use these sub-bands as frequency-domain cues for selective fusion.

We therefore adopt a stage-specific frequency-band selection strategy:
\begin{equation}
\begin{aligned}
S_1 &= \{LH_1,HL_1,HH_1\},&
S_2 &= \{LH_2,HL_2,HH_2\},\\
S_3 &= \{LL_3,LH_3,HL_3\},&
S_4 &= \{LL_4\}.
\end{aligned}
\end{equation}
Accordingly, shallow stages retain high-frequency bands for boundary and detail compensation, intermediate stages combine structural and contour cues, and deep stages keep only the low-frequency band to avoid high-frequency interference. Since DWT sub-bands have lower spatial resolution than the original encoded features, the selected bands are aligned to the spatial size of $C_i$ via bilinear interpolation, denoted as $\mathcal{R}_i(\cdot)$.

For shallow stages $i=1,2$, $HH_i$ captures strong local variations but may also introduce unstable high-frequency responses. Therefore, we generate a high-frequency reliability gate from the original encoded feature and the aligned $HH_i$ component:
\begin{equation}
\begin{aligned}
\bar{HH}_i &= \mathcal{R}_i(HH_i),\\
G_i^{HH} &= \sigma\!\left(\psi_i\left([C_i,\bar{HH}_i]\right)\right),\\
HH_i^g &= G_i^{HH}\odot\bar{HH}_i,\quad i=1,2.
\end{aligned}
\end{equation}
where $\psi_i(\cdot)$ is a lightweight mapping function, $\sigma(\cdot)$ is the sigmoid activation, and $\odot$ denotes element-wise multiplication. The gated component $HH_i^g$ replaces $HH_i$ in shallow frequency fusion.

The selected bands at each stage are spatially aligned and channel-wise concatenated to form the band-fusion input $B_i$, where the gated component $HH_i^g$ is used for $i=1,2$ and the selected bands are directly used for $i=3,4$. A lightweight fusion function $\Phi_i(\cdot)$ maps $B_i$ to the frequency-enhanced feature $F_i$ with the same channel dimension as $C_i$.

To regulate the spatial injection of shallow frequency enhancement, GeoCueFormer generates a geometric spatial gate from the precomputed depth prior $D$ and reinjects the enhanced features into the original encoded features through a residual connection:
\begin{equation}
\begin{aligned}
M_i &= \sigma\!\left(\eta_i(D_i)\right),\quad
\widehat{F}_i=M_i\odot F_i,\quad i=1,2,\\
C'_i &=
\begin{cases}
C_i+\alpha_i\widehat{F}_i, & i=1,2,\\
C_i+\alpha_iF_i, & i=3,4.
\end{cases}
\end{aligned}
\end{equation}
where $D_i=\operatorname{Resize}_{\mathrm{bilinear}}(D,H_i,W_i)$, $\eta_i(\cdot)$ denotes the depth-gating function, and $\alpha_i$ is a learnable residual scaling factor initialized to 0.1. This enables geometry-aware shallow frequency enhancement while preserving the original hierarchical representations.

In implementation, the selected wavelet bands are concatenated along the channel dimension and fused by a $1\times1$ Conv-BN-ReLU block, which projects them back to the original channel dimension of each encoding stage. The depth gate uses a $3\times3$ convolution followed by ReLU, a $1\times1$ convolution, and sigmoid activation, with 8 hidden channels.

\subsection{Prediction-Cued Dual-Stage Decoder}

\begin{figure}[t]
    \centering
    \includegraphics[width=\textwidth]{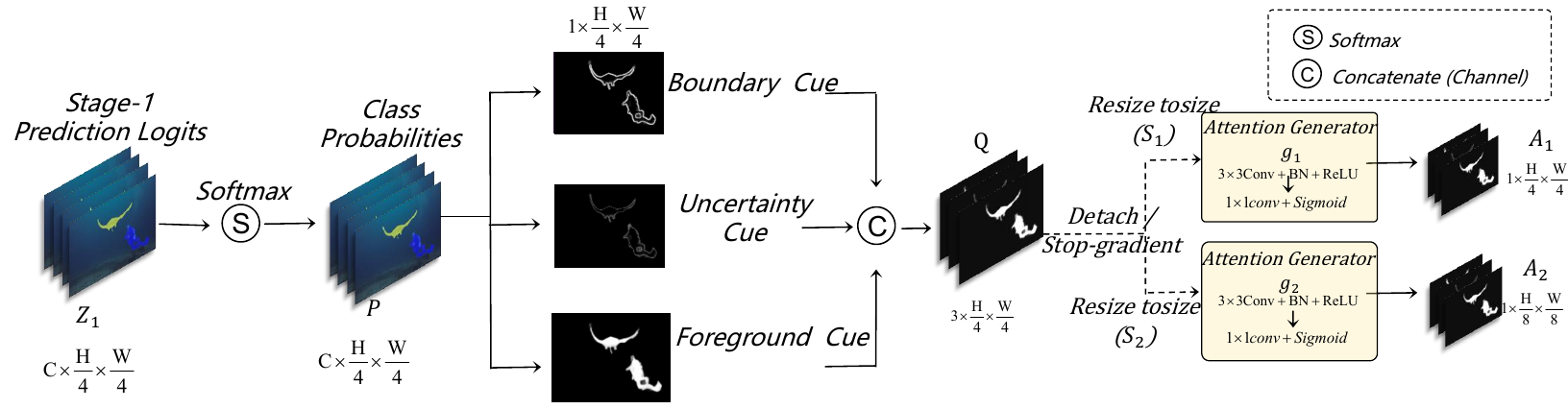}
    \caption{Prediction-Cued Dual-Stage Decoder.}
    \label{fig:dual_decoder}
\end{figure}

To exploit the spatial discriminative information in the initial prediction, PCDD first fuses the enhanced multi-scale features to produce a coarse segmentation prediction. As shown in Fig.~\ref{fig:dual_decoder}, given the enhanced features $\{C'_i\}_{i=1}^{4}$ from Sec.~3.1, the first-stage decoder $H_1(\cdot)$ outputs coarse segmentation logits and the corresponding class probability map:
\begin{equation}
Z_1=H_1(C'_1,C'_2,C'_3,C'_4),\quad
P^{(1)}=\operatorname{Softmax}(Z_1).
\end{equation}
where $P^{(1)}$ denotes the first-stage class probability map. It is not used as the final inference output, but is used to construct prediction cues, while $Z_1$ serves only as an auxiliary supervision term during training.

Both decoders adopt SegFormerHead with input channels $[32,64,160,256]$, decoder embedding dimension 256, intermediate channel dimension 128, dropout ratio 0.1, and 6 output classes.

Based on $P^{(1)}$, we construct the boundary cue $B_p$, uncertainty cue $U_p$, and foreground cue $F_p$ to describe local probability variations, low-confidence regions, and non-background responses, respectively. The three cues are concatenated along the channel dimension to form the prediction cue map $Q$:
\begin{equation}
\begin{aligned}
B_p &= \frac{1}{K}\sum_{k=1}^{K}
\left(|\nabla_x P^{(1)}_k|+|\nabla_y P^{(1)}_k|\right),\\
U_p &= 1-\max_k P^{(1)}_k,\quad
F_p=1-P^{(1)}_{\mathrm{bg}},\\
Q &= \operatorname{SG}\!\left(\operatorname{Concat}(B_p,U_p,F_p)\right).
\end{aligned}
\end{equation}
where $K$ is the number of classes, $P^{(1)}_{\mathrm{bg}}$ denotes the background probability, and $\operatorname{SG}(\cdot)$ denotes stop-gradient operation. Thus, $Q$ serves solely as a second-stage spatial modulation signal, preventing gradients from propagating back to the first-stage prediction branch through the cues. The prediction cue map $Q$ has three channels corresponding to boundary, uncertainty, and foreground cues.

The prediction cue map $Q$ modulates only the first two shallow high-resolution features. After being aligned to each feature size, $Q$ is mapped to the spatial weight $A_j$ by a lightweight function, and the modulated feature is given by:
\begin{equation}
\widetilde{C}_j=C'_j\odot(1+\lambda_jA_j),\quad j=1,2.
\end{equation}
where $\lambda_j$ is a constrained learnable scaling factor controlling the cue modulation strength. Deep features are kept unchanged, i.e., $\widetilde{C}_i=C'_i,\ i=3,4$.

The cue-to-weight mapping uses a $3\times3$ Conv-BN-ReLU block followed by a $1\times1$ convolution and sigmoid activation, with 16 hidden channels, producing a one-channel spatial modulation weight $A_j$ for each shallow feature.

The second-stage decoder $H_2$ re-fuses the modulated multi-scale features and outputs the final logits $Z_2$ for inference. During training, the second-stage output is supervised as the main prediction, while the first-stage output is used for auxiliary supervision:
\begin{equation}
\mathcal{L}
=
\mathcal{L}_{\mathrm{CE}}(\widetilde{Z}_2,Y)
+
0.4\,\mathcal{L}_{\mathrm{CE}}(\widetilde{Z}_1,Y),
\end{equation}
where $\widetilde{Z}_1$ and $\widetilde{Z}_2$ denote the logits resized to the ground-truth label size, and $Y$ is the semantic segmentation ground truth.

\section{Experiments}

\subsection{Experimental Settings}

We evaluate GeoCueFormer on SUIM~\cite{islam2020suim} and DUT-USEG (DUT)~\cite{ma2022dutuseg} using mean Intersection over Union (mIoU), Params, and GFLOPs. Results of compared methods are collected from their original papers or published comparison tables. GeoCueFormer uses a precomputed monocular depth prior generated offline from the input RGB image, and the reported complexity excludes the offline depth estimator. Unless otherwise specified, additional analyses are conducted on SUIM.

\noindent\textbf{Implementation details.}
Experiments are implemented in MMSegmentation. We initialize the MiT-B0 backbone with ImageNet-pretrained weights and train with AdamW, using a base learning rate of $6\times10^{-5}$, betas of $(0.9,0.999)$, and weight decay of 0.01. We use a poly schedule with 1500-iteration linear warm-up, power 1.0, and minimum learning rate 0. Training images are randomly resized within 0.5--2.0, cropped to $480\times640$, horizontally flipped with probability 0.5, and photometrically distorted. The batch size is 8, with 160k iterations on SUIM and 40k on DUT. Both segmentation heads use cross-entropy loss, and the first-stage PCDD loss is weighted by 0.4. Evaluation uses single-scale testing without flip augmentation.

\subsection{Comparison with State-of-the-Art Methods}

Table~\ref{tab:suim_comparison} reports the SUIM comparison. GeoCueFormer achieves 82.23\% mIoU with 4.28M parameters and 13.58 GFLOPs, providing a favorable accuracy-complexity trade-off and outperforming larger models such as Swin Transformer, UperNet, and Mask2Former.

RMP-Net~\cite{chen2022rmpnet} reports 84.52\% mIoU on SUIM, but uses 78.449M parameters and 202.46G GFLOPs, about 18.3$\times$ and 14.9$\times$ those of GeoCueFormer. WaterBiSeg-Net~\cite{zhang2024waterbiseg} reports 85.7\% mIoU under a five-class marine-debris protocol, which differs from our six-class setting. We therefore focus on methods with consistent category protocols and comparable evaluation settings.

\begin{table}[p]
\centering
\caption{Comparison with state-of-the-art methods on the SUIM dataset. N/R denotes unavailable results.}
\label{tab:suim_comparison}
\small
\setlength{\tabcolsep}{4pt}
\renewcommand{\arraystretch}{0.95}
\begin{tabular}{lccc}
\hline
Method & GFLOPs$\downarrow$ & Params$\downarrow$ & mIoU (\%)$\uparrow$ \\
\hline
GeoCueFormer (Ours) & 13.58 & 4.28M & \textbf{82.23} \\
UWSegFormer~\cite{zuo2025uwsegformer} & 3.21 & 21.78M & \underline{82.12} \\
LVT~\cite{yang2022lvt} & 10.51 & 3.83M & 80.72 \\
Swin Transformer~\cite{liu2021swin} & 276 & 58.94M & 80.70 \\
SegFormer~\cite{xie2021segformer} & 7.94 & 3.72M & 80.57 \\
MSNet~\cite{liu2026msnet} & N/R & 6.2M & 79.83 \\
Swin-EMA-FPN SegFormer~\cite{chen2025improvedsegformer} & 26.51 & 91.86M & 77.00 \\
DDRNet-s~\cite{hong2021ddrnet} & 5.35 & 5.7M & 73.44 \\
UperNet~\cite{xiao2018upernet} & 44.48 & 122.88M & 72.70 \\
Mask2Former~\cite{cheng2022mask2former} & 226.63 & 44.63M & 72.55 \\
UISS-Net / USS-NET~\cite{he2024uissnet} & 342 & 75.42M & 72.09 \\
PIDNet-s~\cite{xu2023pidnet} & 6.96 & 7.6M & 71.46 \\
MaskFormer~\cite{cheng2021maskformer} & 53.22 & 41.27M & 71.07 \\
DeepLabV3+~\cite{chen2018encoder} & 80.06 & 41.22M & 71.03 \\
KNet~\cite{zhang2021knet} & 37.83 & 60.34M & 70.54 \\
PSPNet~\cite{zhao2017pyramid} & 61.68 & 46.61M & 69.76 \\
ISANet~\cite{huang2019isanet} & 28.09 & 35.34M & 68.76 \\
\hline
\end{tabular}
\end{table}

Table~\ref{tab:dut_comparison} presents the DUT comparison. GeoCueFormer achieves 73.04\% mIoU, the best reported performance among existing published methods on DUT, showing consistent gains under weak foreground-background contrast and ambiguous boundaries.

\begin{table}[p]
\caption{Comparison with state-of-the-art methods on the DUT dataset. N/R denotes unavailable results.}
\label{tab:dut_comparison}
\centering
\small
\setlength{\tabcolsep}{4pt}
\renewcommand{\arraystretch}{0.95}
\begin{tabular}{lccc}
\hline
Method & GFLOPs ($\downarrow$) & Params ($\downarrow$) & mIoU (\%) \\
\hline
GeoCueFormer (Ours) & 13.58 & 4.28M & \textbf{73.04} \\
UWSegFormer~\cite{zuo2025uwsegformer} & 3.21 & 21.78M & \underline{71.41} \\
Lightweight DeepLabv3+~\cite{chen2018encoder} & 39.612 & 6.628M & 71.18 \\ % TODO: add the source of the exact lightweight variant.
SegFormer~\cite{xie2021segformer} & 7.94 & 3.72M & 70.81 \\
UHRS-Net~\cite{zhou2025uhrsnet} & N/R & N/R & 70.09 \\
HRNetv2~\cite{wang2021deep} & 90.972 & 29.540M & 70.09 \\
Swin Transformer~\cite{liu2021swin} & 276 & 58.94M & 69.35 \\
DDRNet~\cite{hong2021ddrnet} & 5.35 & 5.70M & 69.15 \\
PSPNet~\cite{zhao2017pyramid} & 61.68 & 46.708M & 68.51 \\
UISS-Net~\cite{he2024uissnet} & 342 & N/R & 67.87 \\
PIDNet~\cite{xu2023pidnet} & 6.96 & 7.60M & 67.79 \\
DeepLabv3+~\cite{chen2018encoder} & 80.06 & 5.814M & 67.63 \\
U-Net~\cite{ronneberger2015unet} & 238 & 43.933M & 67.19 \\
FCN~\cite{long2015fully} & 46.11 & 9.71M & 62.75 \\
SeaFormer~\cite{wan2023seaformer} & 7.55 & 14.00M & 57.54 \\
\hline
\end{tabular}
\end{table}

\subsection{Qualitative Comparison}

\begin{figure}
    \centering
    \includegraphics[width=0.95\textwidth]{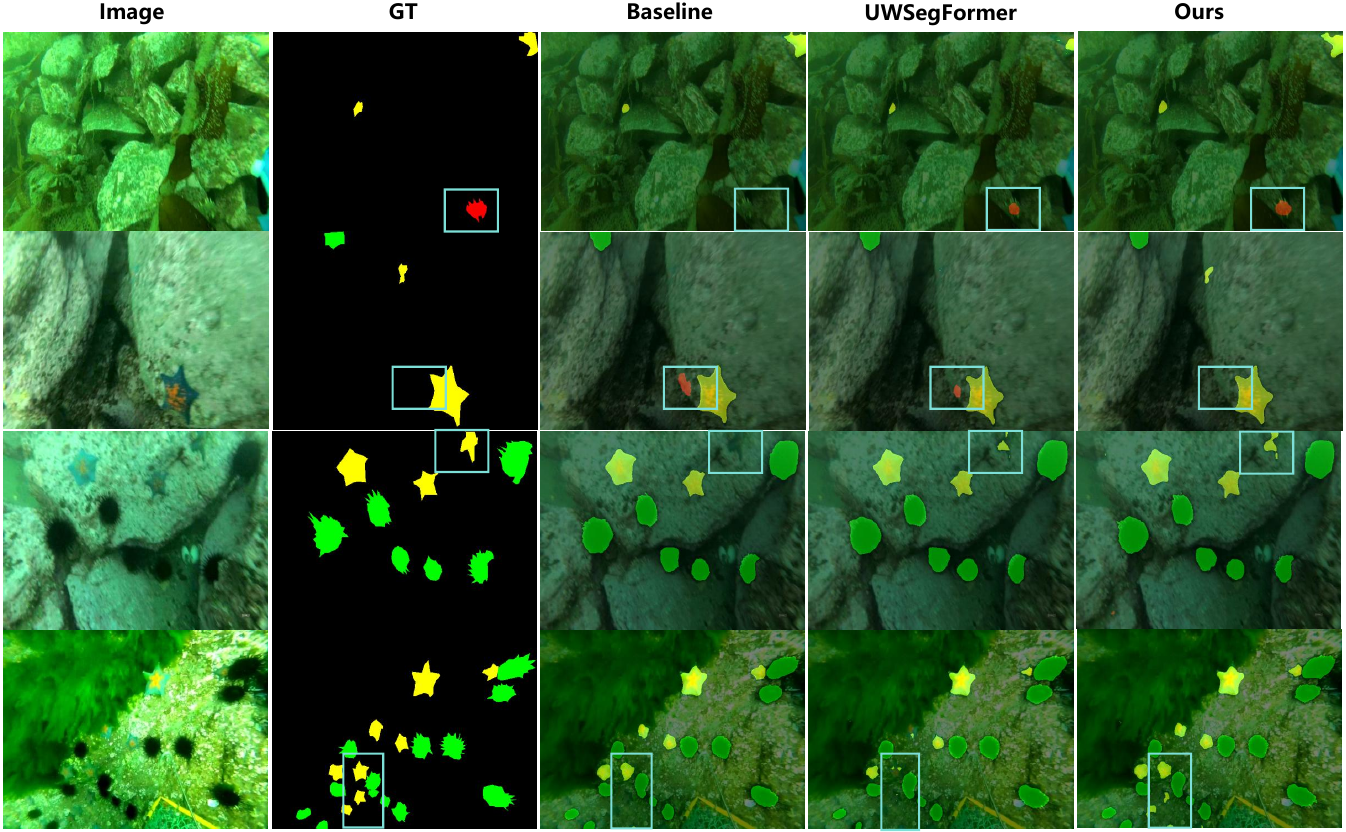}
    \caption{Qualitative comparison of segmentation results. Cyan boxes highlight challenging regions with low contrast, small objects, or ambiguous boundaries.} \label{fig:qualitative_comparison}
\end{figure}
Fig.~\ref{fig:qualitative_comparison} compares GeoCueFormer with SegFormer and UWSegFormer. GeoCueFormer produces more complete regions and clearer boundaries in challenging underwater scenes.
\subsection{Ablation Study}

Table~\ref{tab:main_ablation} evaluates each component on SUIM. Wavelet enhancement and PCDD individually improve SegFormer, but their direct combination without geometric guidance does not bring further gains, indicating that unconstrained frequency cues may disturb prediction-cued refinement. By embedding geometric guidance into stage-specific wavelet enhancement, GSWE improves the reliability of frequency representation, and the full model achieves the best performance, confirming the complementarity between GSWE and PCDD. We omit a “geometry guidance only” variant because geometric guidance is used to constrain the shallow wavelet branch rather than as an independent module.

\begin{table}[!htbp]
\caption{Ablation study of the main components on SUIM.}
\label{tab:main_ablation}
\centering
\small
\setlength{\tabcolsep}{4pt}
\begin{tabular}{lcccc}
\hline
Variant & Wavelet & Depth & PCDD & mIoU (\%) \\
\hline
SegFormer Baseline & $\times$ & $\times$ & $\times$ & 80.57 \\
+ Wavelet & $\checkmark$ & $\times$ & $\times$ & 80.98 \\
+ Two-stage Decoder & $\times$ & $\times$ & $\checkmark$ & 81.42 \\
+ Wavelet + Two-stage Decoder & $\checkmark$ & $\times$ & $\checkmark$ & 81.21 \\
+ Wavelet + Depth & $\checkmark$ & $\checkmark$ & $\times$ & 81.58 \\
Full Model & $\checkmark$ & $\checkmark$ & $\checkmark$ & 82.23 \\
\hline
\end{tabular}
\end{table}

\subsection{Analysis of Frequency Strategy and Depth Injection}

Table~\ref{tab:frequency_strategy} compares frequency-band strategies. Full-band fusion improves the baseline but remains below the stage-specific design, indicating that indiscriminate band fusion may introduce redundant textures and unstable high-frequency responses.

\begin{table}[!htbp]
\caption{Analysis of frequency-band selection strategy on SUIM.}
\label{tab:frequency_strategy}
\centering
\small
\setlength{\tabcolsep}{5pt}
\begin{tabular}{lccc}
\hline
Variant & Frequency strategy & Decoder & mIoU (\%) \\
\hline
Baseline & - & Single & 80.57 \\
Full-stage Full-band & LL/LH/HL/HH & Two-stage & 80.91 \\
GeoCueFormer & Specific / Selective & Two-stage & 82.23 \\
\hline
\end{tabular}
\end{table}
Table~\ref{tab:depth_position} compares depth-injection positions. Shallow-stage guidance performs best, while all-stage gating reduces mIoU, suggesting that excessive deep geometric modulation can disturb semantic representations.
\begin{table}[H]
\caption{Analysis of depth-guidance injection position on SUIM.}
\label{tab:depth_position}
\centering
\small
\setlength{\tabcolsep}{5pt}
\begin{tabular}{llc}
\hline
Setting & Depth-guidance stages & mIoU (\%) \\
\hline
 GeoCueFormer & Stage 1 + Stage 2 & 82.23 \\
 Depth-All & Stage 1 + Stage 2 + Stage 3 + Stage 4 & 80.06 \\
\hline
\end{tabular}
\end{table}

\subsection{Hyperparameter Study on Residual Scale Initialization}
Table~\ref{tab:alpha_init} studies the residual scale initialization. The best result is obtained at $\alpha_{\text{init}}=0.1$, balancing backbone semantics with frequency and geometry enhancement, while a larger value may disturb pretrained representations.

\begin{table}[H]
\caption{Analysis of the initial residual scale $\alpha_{\text{init}}$ on SUIM.}
\label{tab:alpha_init}
\centering
\small
\setlength{\tabcolsep}{8pt}
\renewcommand{\arraystretch}{0.95}
\begin{tabular}{lcc}
\hline
Setting & $\alpha_{\text{init}}$ & mIoU (\%) \\
\hline
GeoCueFormer & 0.01 & 80.40 \\
GeoCueFormer & 0.05 & 81.44 \\
GeoCueFormer & 0.10 & \textbf{82.23} \\
GeoCueFormer & 0.20 & 81.09 \\
\hline
\end{tabular}
\end{table}

\section{Conclusion}

We proposed GeoCueFormer, a lightweight underwater segmentation framework that integrates geometry-guided frequency representation and prediction-cued refinement. By combining stage-specific wavelet enhancement, depth-based reliability gating, and a dual-stage decoder, GeoCueFormer improves degraded boundary and low-contrast object perception with a small parameter footprint and moderate computational cost. Experiments on SUIM and DUT show that GeoCueFormer achieves SOTA-level performance under comparable model complexity and standard benchmark settings, while maintaining a favorable accuracy-complexity trade-off. Future work will explore learnable frequency-band selection for more adaptive underwater representation.

\clearpage
\renewcommand{\refname}{References}

\end{document}